%% file: IV2019.tex
\documentclass[letterpaper, 10pt, conference]{ieeeconf}

\pdfoutput=1
\IEEEoverridecommandlockouts        	
\usepackage[utf8]{inputenc}
\usepackage[T1]{fontenc}

\usepackage[pdftex]{graphicx}
\usepackage{epstopdf}
\usepackage{amsmath} 
\usepackage{amssymb} 
\usepackage{nicefrac}
\usepackage{booktabs}
\usepackage{multirow}
\usepackage{siunitx}
\usepackage{microtype}
\usepackage[vlined,ruled]{algorithm2e}
\usepackage[table,xcdraw]{xcolor}
\usepackage{multicol}
\usepackage{accents}
\usepackage{flushend}
\usepackage{fancyhdr}

\makeatletter
\let\MYcaption\@makecaption
\makeatother

\usepackage[font=footnotesize]{subcaption}

\makeatletter
\let\@makecaption\MYcaption
\makeatother

\makeatletter
\let\NAT@parse\undefined
\makeatother
\usepackage[numbers,sort,square]{natbib}

\usepackage{hyperref}

\renewcommand{\tableautorefname}{Tab.}
\renewcommand{\figureautorefname}{Fig.}
\renewcommand{\sectionautorefname}{Sec.}
\renewcommand{\subsectionautorefname}{Sec.}
\renewcommand{\algorithmautorefname}{Alg.}
\def\equationautorefname~#1\null{(#1)\null}

\graphicspath{{./img/}}

\newcommand{\bestResult}[1]{{\color[HTML]{32CB00} \textbf{#1}}}

\input{math.tex}

\title{\LARGE \bf Incrementally learned Mixture Models for GNSS Localization}

\author{\authorblockN{Tim Pfeifer and Peter Protzel}%
	\authorblockA{Dept.~of Electrical Engineering and Information Technology\\
		TU Chemnitz, Germany\\
		Email: \{firstname.lastname\}@etit.tu-chemnitz.de}%
	\thanks{The project is funded by the "Bundesministerium f\"ur Wirtschaft und Energie" (German Federal Ministry for Economic Affairs and Energy).}%
}

\begin{document}
	
	\maketitle
	\thispagestyle{fancy}
	\fancyhf{}
	\fancyhead[OL]{ 
		\footnotesize
		Proc. of IEEE Intelligent Vehicles Symposium (IV), 2019, Paris, France. DOI: 10.1109/IVS.2019.8813847\\
		\tiny
		\copyright 2019 IEEE. Personal use of this material is permitted. Permission from IEEE must be obtained for all other uses, in any current or future media, including
		reprinting/republishing this material for advertising or promotional purposes, creating new collective works, for resale or redistribution to servers or lists, or reuse of any copyrighted component of this work in other works.}
	
	\addtolength{\headheight}{\baselineskip}
	
	\begin{abstract}
		GNSS localization is an important part of today's autonomous systems, although it suffers from non-Gaussian errors caused by non-line-of-sight effects.
		Recent methods are able to mitigate these effects by including the corresponding distributions in the sensor fusion algorithm.
		However, these approaches require prior knowledge about the sensor's distribution, which is often not available.
		
		We introduce a novel sensor fusion algorithm based on variational Bayesian inference, that is able to approximate the true distribution with a Gaussian mixture model and to learn its parametrization online.
		The proposed Incremental Variational Mixture algorithm automatically adapts the number of mixture components to the complexity of the measurement's error distribution.
		We compare the proposed algorithm against current state-of-the-art approaches using a collection of open access real world datasets and demonstrate its superior localization accuracy.
	\end{abstract}
	
	\section{Introduction}
	
	A reliable localization is crucial for autonomous systems like self-driving cars or mobile robots.
	While global navigation satellite systems (GNSS) can provide this information for open-field scenarios, their accuracy and consistency is decreased in urban areas.
	The reason are signal reflections on tall buildings, which cause non-line-of-sight (NLOS) measurements.
	Recent publications like \cite{Hsu2017} demonstrated the non-Gaussian characteristic of the resulting error distributions.
	Therefore, a lot of effort has already been invested to make GNSS more robust against NLOS effects \cite{Bressler2016, Zhu2018}.
	
	One approach to reduce the influence of these errors, is to incorporate knowledge about their true distribution into the state estimation process.
	With robust factor graph optimization, there already is a class of algorithms that allows to include such distributions either in a predefined \cite{Olson2012, Rosen2013} or self-tuning \cite{Pfeifer2018,Pfeifer2019,Watson2018a} way.
	To describe the non-Gaussian errors, Gaussian mixture models (GMM) are the established state-of-the-art.
	However, existing approaches are not able to fully adapt the GMM during the online estimation process.
	Therefore, they do not address applications where the error characteristic changes significantly over time.
	For GNSS this can easily happen, since the NLOS error depends on the receiver's surrounding.
	
	Based on our recently published Adaptive Mixture algorithm \cite{Pfeifer2019}, we want to introduce a novel approach that combines variational Bayesian inference \cite{Corduneanu2001} with non-linear graph optimization to achieve an even better robustness and accuracy.
	Due to an incremental construction of the GMM, we preserve the real-time capability of the original approach and eliminate existing drawbacks.
	
	Since the proposed algorithm is the result of an incremental research process, we want to give an overview over prior work in \autoref{sec:prior}.
	How GNSS localization and Gaussian mixtures are connected to graph based optimization is explained in \autoref{sec:GNSS}, while \autoref{sec:Mixture} gives an overview about the algorithms that are used to estimate Gaussian mixtures from data.
	Then we combine these techniques to the proposed Incremental Variational Mixture approach in \autoref{sec:Complexity} before we evaluate it in \autoref{sec:Eval} using several real world GNSS datasets and finally conclude. 
	
	\begin{figure}
		\centering
		\includegraphics[width=\linewidth]{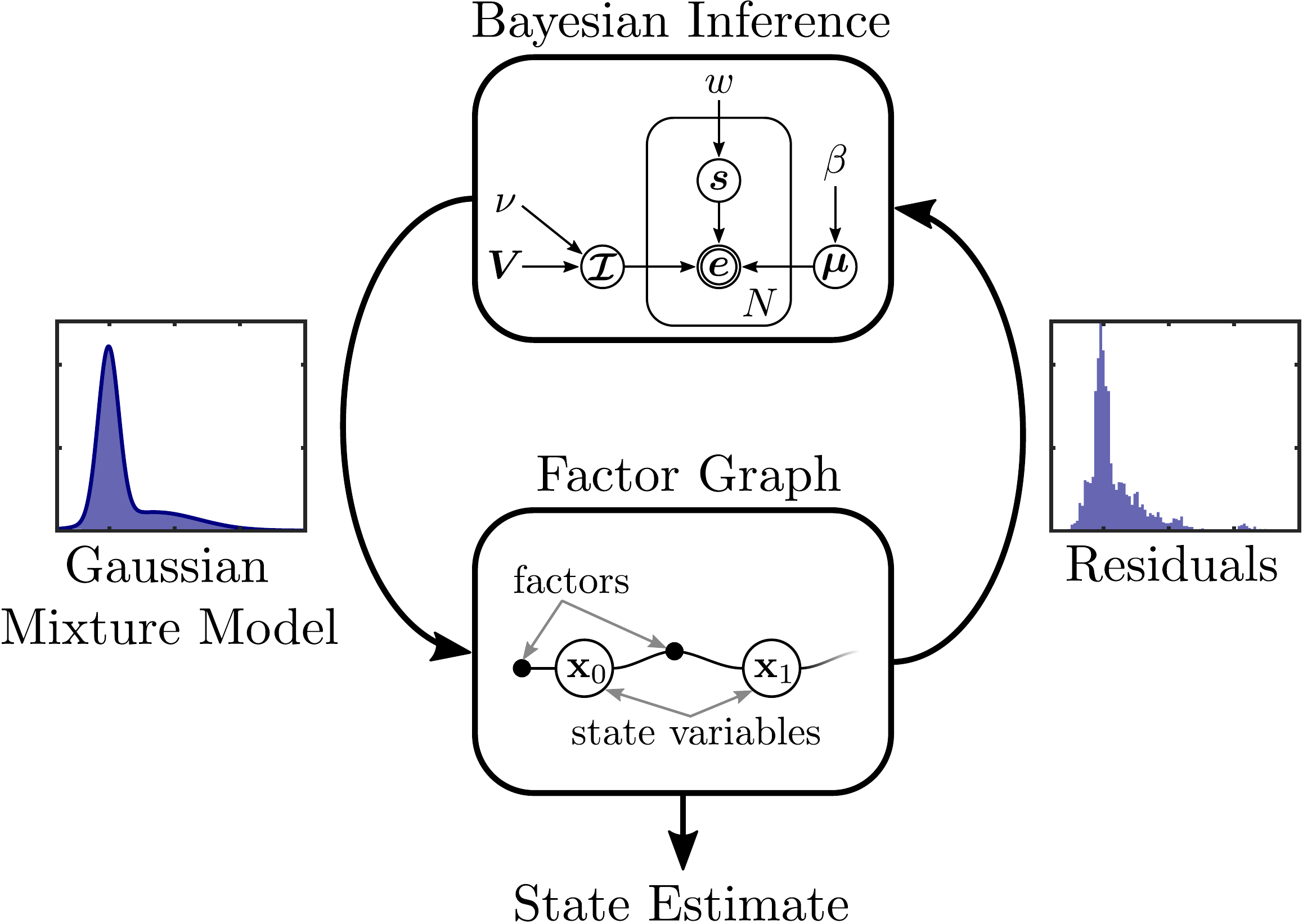}
		\caption
		{
			The combination of Bayesian inference and factor graph optimization allows robust state estimation even if the true error distribution is unknown.
			Our approach incrementally learns the non-Gaussian error distribution simultaneously to the state estimation process.
		}
		\label{fig:Concept}
	\end{figure}
	
	\section{Prior Work}\label{sec:prior}
	Several recently published algorithms try to solve the problem of non-Gaussian sensor fusion in different ways.
	From our point of view, they can be separated by two properties of their error models: If they are using static or dynamic ones and whether the models are Gaussian or not.
	We summarize existing work in \autoref{tab:prior-work}, categorized by the error model's properties.
	For example, the standard least squares approach \cite{Dellaert2017} is a static Gaussian method since it uses a Gaussian error model that is predefined.
	
	\begin{table}[tbph]
		\centering
		\caption{Robust Factor Graph Algorithms categorized regarding the applied Error Model}
		\label{tab:prior-work}
		\begin{tabular}{@{}l|c|c@{}}
			\toprule
			\textbf{Error Model} & \multicolumn{1}{c|}{\textbf{Static}} & \multicolumn{1}{c}{\textbf{Dynamic}} \\ \midrule
			\textbf{Gaussian} & Standard Factor Graph & \begin{tabular}[c]{@{}c@{}}SC \cite{Suenderhauf2012a}\\ DCE \cite{Pfeifer2017}\end{tabular} \\ \midrule
			\textbf{Non-Gaussian} & 
			\begin{tabular}[c]{@{}c@{}}
				DCS \cite{Agarwal2013} \\
				cDCE \cite{Pfeifer2017} 	\\
				Max-Mix. \cite{Olson2012} 	\\
				Sum-Mix. \cite{Rosen2013}
			\end{tabular} & 
			\begin{tabular}[c]{@{}c@{}}
				Self-tuning M-Est. \cite{Agamennoni2015}	\\ 
				Self-tuning Mix. \cite{Pfeifer2018}			\\
				Adaptive Mix. \cite{Pfeifer2019}		\\
				Batch Cov. Est. \cite{Watson2018a}
			\end{tabular} \\ 
			\bottomrule
		\end{tabular}
	\end{table}
	
	With Switchable Constraints (SC) \cite{Suenderhauf2012a} and it's modification Dynamic Covariance Estimation (DCE) \cite{Pfeifer2017} exist two algorithms that implement Gaussian but dynamic error models.
	While SC introduces tunable weights for each Gaussian, DCE estimates the covariance of the distribution.
	Both are able to handle Gaussian distributions with outliers but tend to fail for heavily multimodal distributions that can occur when GNSS is used in urban canyons.
	
	A common robust method, not only in optimization, is the application of predefined non-Gaussian weight functions, so called M-estimators.
	These static non-Gaussian methods include not only the well-known Huber or Tukey estimators but also techniques like Dynamic Covariance Scaling (DCS) \cite{Agarwal2013} or the closed form of DCE (cDCE) \cite{Pfeifer2017}.
	Usually they come with a tuning parameter to adjust the trade-off between robustness and well-behaved convergence.
	Beside M-estimators, there are two approaches to include different distributions in the optimization problem.
	Max-Mixtures \cite{Olson2012} allows approximated inference over a Gaussian mixture by replacing the sum-operator by a max-operator.
	The work of Rosen \cite{Rosen2013} includes all kinds of continuous probability distributions including Gaussian mixtures.
	However, all methods with static error models require exact knowledge about the expected error distribution.
	For GNSS in changing environments, this information is often not available. 
	
	Algorithms that apply non-Gaussian error models and estimate their parameter dynamically are a relatively new class of robust solutions to non-linear optimization.
	The authors of \cite{Agamennoni2015} were the first who described an algorithm that is capable to estimate the optimal parameter of a certain group of M-estimators, using adaptive importance sampling.
	Later in \cite{Pfeifer2018}, the parameters of a Gaussian mixture model (GMM) were optimized simultaneously during the estimation process.
	In recent work \cite{Pfeifer2019}, Expectation-Maximization (EM) was applied to estimate the GMM in a more reliable way.
	Also, \citeauthor{Wang2018} introduced a similar algorithm in the field of laser scan matching \cite{Wang2018}.
	They estimate a mixture of exponential power to describe the residuals of the point-to-point registration process.
	The prerequisite for this type of algorithms is the selection of a suitable model respectively its complexity.
	For GNSS applications this can be a limitation, since the complexity (e.g. the number of modes) can change with the environment over time.
	
	\citeauthor{Watson2018a} currently proposed a novel algorithm, based on the infinite Gaussian mixture model, which is adaptive regarding its number of components \cite{Watson2018a}.
	Since the applied Gibbs sampling is not real-time capable, the approach can only be applied to batch problems.
	For the GNSS localization of an autonomous system this is a harsh limitation, since it usually has to be solved in real time. 
	Also, the algorithm applies one common model for the whole dataset, so it cannot address time dependent changes.
	
	While existing algorithms tune a predefined distribution to approximate the estimation error, we want to go one step further and learn its structure and parametrization online.
	Therefore, we want to propose an algorithm that is able to exploit the changing complexity of the error distribution in urban scenarios.
	Compared to the most recent self-tuning mixtures algorithm \cite{Pfeifer2019}, our approach is novel in the following ways:
	\begin{enumerate}
		\item The proposed algorithm incrementally constructs a Gaussian mixture with a (theoretically) unlimited number of components.
		\item We are able to remove unused Gaussian components to preserve real-time capability.
		\item To ensure a numerically stable solution, we apply variational Bayesian inference instead of Expectation-Maximization to estimate the GMM.
	\end{enumerate}
	With this combination of techniques, we propose the first real-time capable sensor fusion algorithm, that is able to learn the complexity and parametrization of the measurement's error distribution over time.
	
	\section{GNSS Localization as Factor Graph}\label{sec:GNSS}
	
	In the following section, we want to give a brief overview how the GNSS problem can be described as factor graph and how non-Gaussian error distributions can be represented.
	Factor graphs, a graphical representation of least squares problems, are wildly used in robotics.
	Therefore, we assume basic knowledge about their theoretical properties and refer readers to \cite{Dellaert2017} for a more detailed introduction.
	
	\subsection{Factor graphs and Least Squares}
	The GNSS localization of a moving system can be described as optimization problem \autoref{eqn:max-like}, where $\Measurement$ is a set of pseudorange and odometry measurements and $\State$ is the set of estimated states including the vehicle's position.
	$\StateOpt$ is the most likely set of states according to the set of measurements.
	By applying the Bayes theorem, it can be written as maximum a posteriori (MAP) problem \autoref{eqn:map} with $\StateEst$ as MAP estimator of the true state variables.
	\begin{equation} 
	\StateOpt =\argmax_{\State} \prob(\State|\Measurement)
	\label{eqn:max-like}
	\end{equation}
	\begin{equation}
	\StateEst =\argmax_{\State} \prob(\Measurement|\State) \prob(\State)
	\label{eqn:map}
	\end{equation}
	
\begin{figure}
		\centering
		\includegraphics[width=\linewidth]{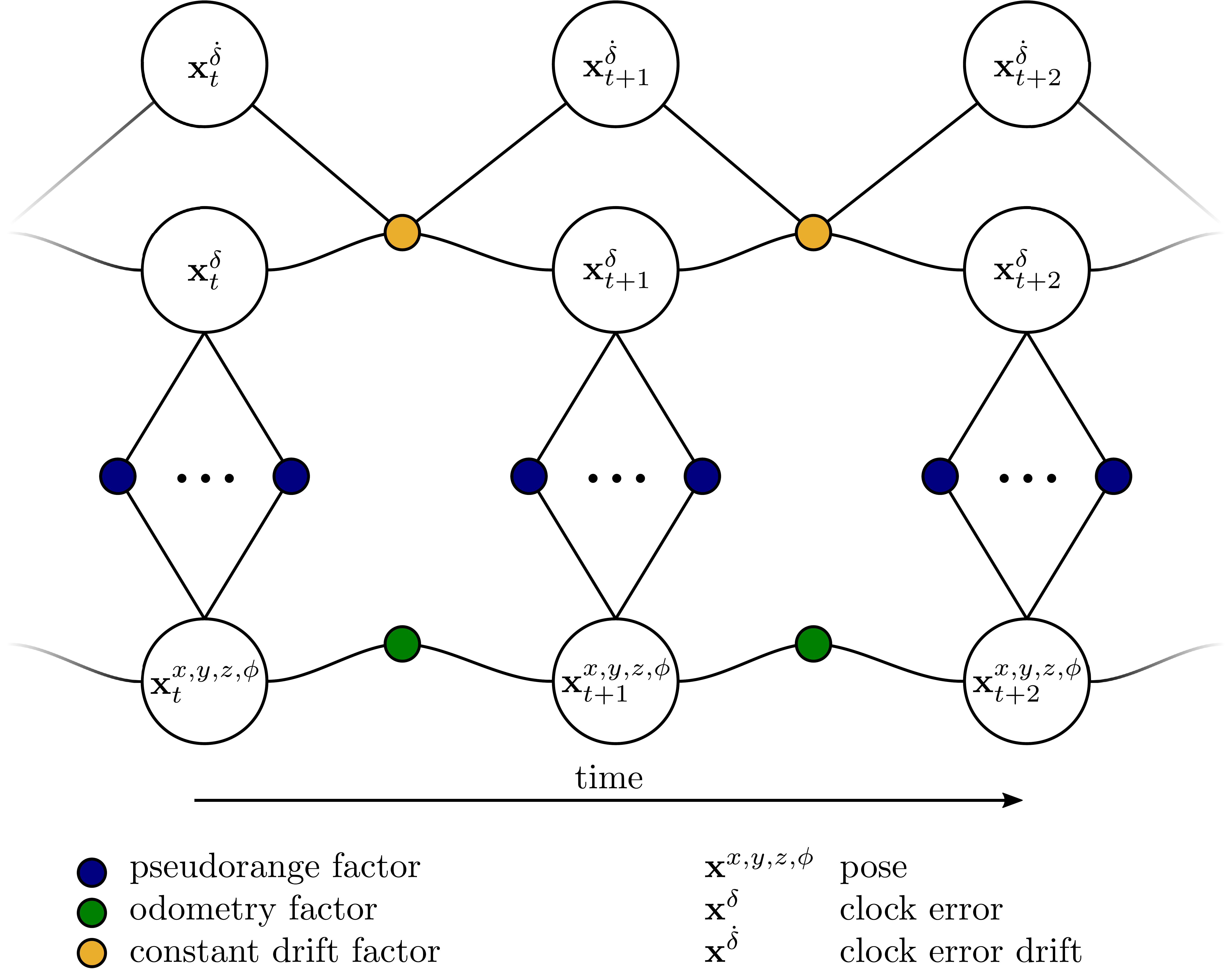}
		\caption{
			Factor graph model of the GNSS localization problem.
			Big circles represent the estimated state variables and small colored dots the error functions which are factors of \autoref{eqn:posterior}.
			The pseudorange factors suffer from non-Gaussian NLOS errors, therefore robust error models have to be applied.}
		\label{fig:FG-GNSS}
	\end{figure}

	At each time step $t$, one odometry measurement $\measurementOdom$ and multiple pseudorange measurements $\measurementPseudorange$, each corresponding to the $i\text{th}$ satellite, are available.
	The set of state variables can be separated in two subsets that consist of the 3D pose $\statePose$ and the GNSS specific clock error state $\stateCCED$.
	The Cartesian coordinates $x,y,z$ are in the earth-centered-earth-fixed (ECEF) frame, while $\phi$ denotes the rotation around the vehicles upright axis.
	The clock error $\delta$ and its derivation $\dot{\delta}$ are required for the constant clock error drift model (CCED).
	By assuming uninformative priors, $\prob(\State)$ can be omitted and the posterior likelihood is expanded to \autoref{eqn:posterior}.
	A detailed explanation of these error terms can be found in \cite{Pfeifer2018}.
	\autoref{fig:FG-GNSS} shows a small example of the corresponding factor graph.
	\begin{equation}
	\begin{split}
	\prob(\Measurement|\State) \propto 
	&\prod_t
	\left[ 
	\prod_{i}
	\prob(\measurementPseudorange|\statePseudorange)
	\right]\\
	\cdot
	&\prod_t
	\prob(\measurementOdom|\statePoseFuture,\statePose)\\
	\cdot
	&\prod_t
	\prob(\stateCCEDFuture|\stateCCED)
	\end{split}
	\label{eqn:posterior}
	\end{equation}
	
	Instead of maximizing the likelihood \autoref{eqn:map} directly, $\StateEst$ is usually estimated by minimizing the negative log likelihood:
	\begin{equation}
	\StateEst = \argmin_{\State} \sum_{n} -\ln(\prob(\measurement_n|\state_n))
	\label{eqn:log_like}
	\end{equation}
	One common index $n$ is used for all factors here.
	By defining $\prob(\Measurement|\State)$ as normal distribution $\mathcal{N}(\mean_n,\info_n)$ over the non-linear error function $\error_n = f(\state_n,\measurement_n)$ with mean $\mean_n$ and information matrix $\info_n$, the problem can be formulated as non-linear least squares estimation:
	\begin{equation}
	\StateEst = \argmin_{\State} \sum_{n} \frac{1}{2} \left\| \sqrtinfon \left( \error_n - \mean_n \right) \right\| ^2
	\label{eqn:least_squares}
	\end{equation}
	As already mentioned, the Gaussian assumption does not apply to all GNSS measurements.
	Therefore, a reliable state estimation requires more advanced distributions.
	
	\subsection{Factor Graphs and Gaussian Mixtures}
	Since empirical sensor distribution can be asymmetric or even multimodal, a class of distributions is required that is able to represent these properties.
	For this work we choose GMMs, because they fulfill this condition and have several positive properties regarding their application in autonomous systems \cite{Rosen2013a}.
	To apply them to a least squares problem, we use a solution proposed in \cite{Rosen2013} that allows almost arbitrary non-Gaussian distributions $\prob(\measurement_n|\state_n)$, using the following:
	\begin{equation}
	\StateEst = \argmin_{\State} \sum_{n} 
	\left\| 
	\sqrt
	{
		-\ln \left( 
		\frac{\prob(\measurement_n|\state_n)}{\gamma} 
		\right) 
	}
	\right\| ^2
	\label{eqn:sum_mix}
	\end{equation}
	The normalization constant $\gamma$ ensures a numerically stable solution.
	We reference the approach as Sum-Mixture (SM) in difference to Max-Mixture of \cite{Olson2012}, which only approximates a GMM.
	In this work, a $K\text{-component}$ Gaussian mixture defined by \autoref{eqn:gaussian_mixture} is used to describe the pseudorange error $\errorPseudorange = f(\statePseudorange, \measurementPseudorange)$.
	Each Gaussian component $k$ is scaled with a normalized weight $\weight_k$.
	\begin{equation}
	\errorPseudorange \sim \sum_{k=1}^{K} \weight_k\cdot\mathcal{N}(\mean_k,\info_k) \text{ with } \sum_{k=1}^{K} \weight_k = 1
	\label{eqn:gaussian_mixture}
	\end{equation}
	Since our approach is not limited to the pseudorange error, we use the generic error $\error$ instead of $\errorPseudorange$ for the following equations.
	
	The likelihood of the GMM is defined by:
	\begin{equation}
	\begin{split}
	\prob(\measurement_n|\state_n) =& 
	\sum_{k}{ c_k \cdot \exp \left( -\frac{1}{2} \left\| \sqrtinfok \left( \error - \mean_k \right) \right\| ^2 \right)}\\
	\text{with }c_k =& \weight_k \cdot \det\left( \sqrtinfok \right)
	\end{split}
	\label{eqn:gaussian_mixture_prob}
	\end{equation}
	Therefor, the normalization constant can be set to $\gamma = \sum_{k} c_k$.
	For a more detailed derivation of these equations we want tor refer the reader to the original work \cite{Rosen2013} and to \cite{Pfeifer2019} for its application to GMMs.
	
	\section{Mixture Model Estimation}\label{sec:Mixture}
\begin{figure}
		\centering
		\begin{subfigure}[t]{0.4\linewidth}
			\centering
			\includegraphics[height=3.1cm]{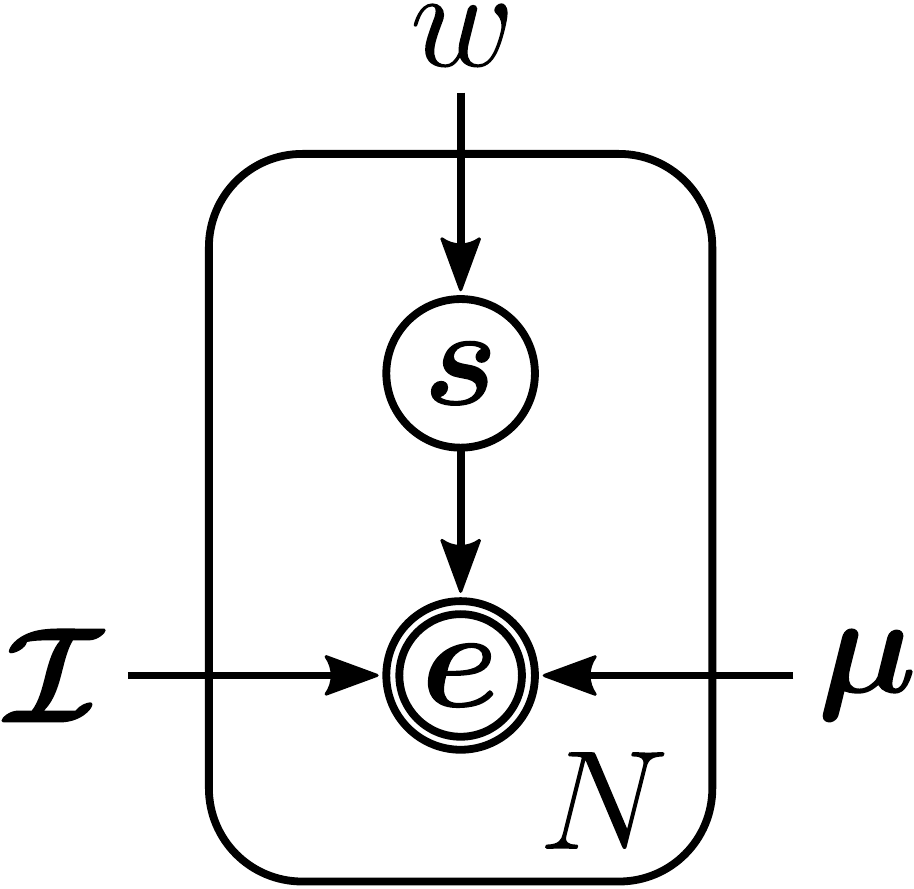}
			\caption{}
			\label{fig:GMM_EM}
		\end{subfigure}
		\hskip 10pt
		\begin{subfigure}[t]{0.52\linewidth}
			\centering
			\includegraphics[height=3.1cm]{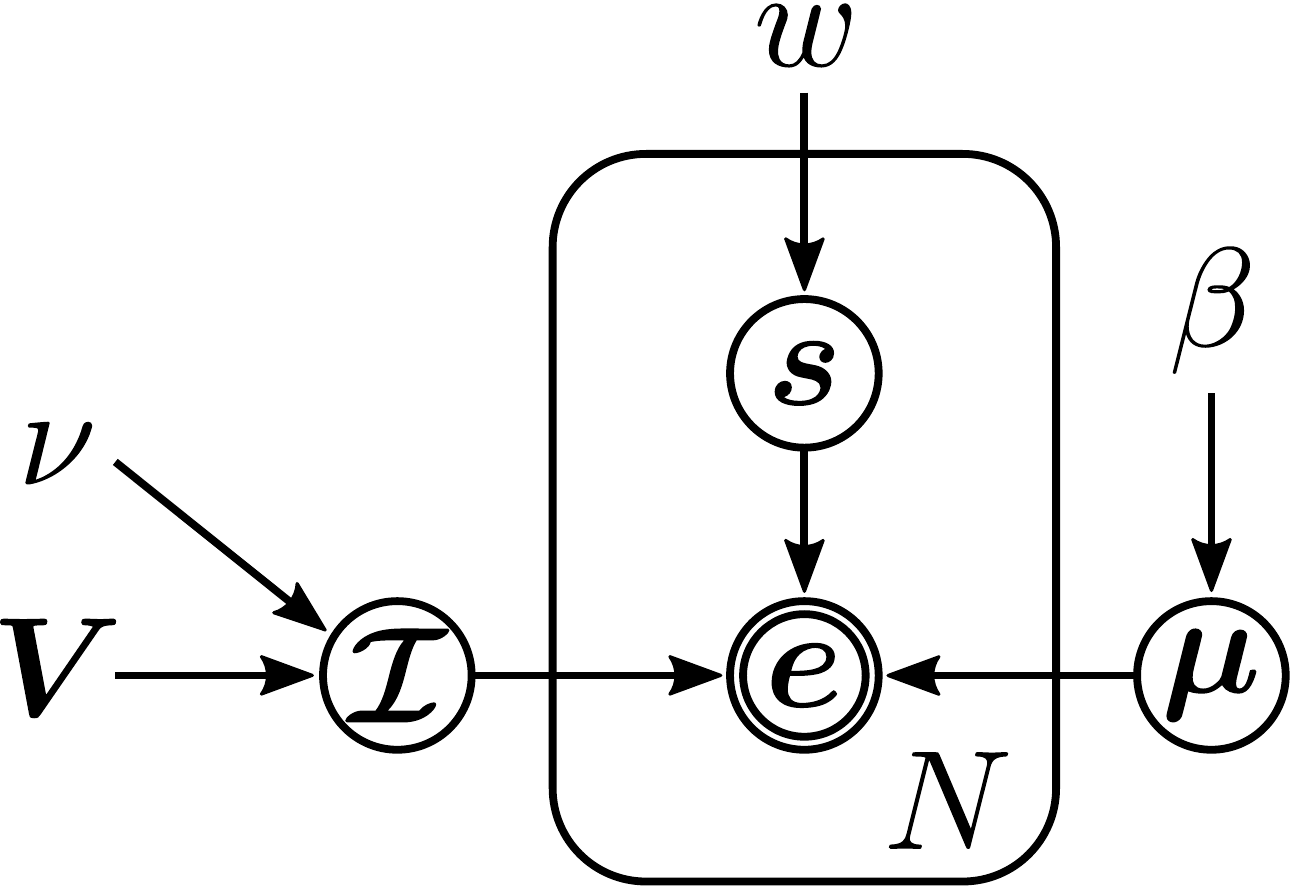}
			\caption{}
			\label{fig:GMM_VBI}
		\end{subfigure}
		\caption
		{
			Directed graphs in plate notation of the maximum-likelihood (a) and the Bayesian (b) interpretation of a Gaussian mixture.
			The circles mark hidden random variables and the double circles the observed ones.
			Unmarked letters are parameters of the corresponding model.
			A legend is provided by \autoref{tab:legend}.
		}
		\label{fig:GMM}
	\end{figure}

	As shown in prior work \cite{Agamennoni2015,Watson2018a,Pfeifer2019}, the residual $\error(\state,\measurement)$ of the estimation problem can be used to get an approximation of the measurement's true distribution.
	To describe the measurement error with a $K$-component Gaussian mixture model, the GMM's parameters $\Params = \left\lbrace \weight,\mean,\info \right\rbrace $ have to be estimated with \autoref{eqn:max-posteriori} from $N$ samples of $\error$.
	\begin{equation}
	\ParamsOpt = \argmax_{\Params} \prob (\Params|\error)
	\label{eqn:max-posteriori}
	\end{equation}
	The set of parameters $\Params$ includes a weight $\weight_k$, mean $\mean_k$ and information matrix $\info_k$ for each component $k$ of $K$.
	Since the probability that measurement $n$ belongs to component $k$ is not known, the corresponding hidden parameter $\boldsymbol{s} = \left\lbrace s_{kn} \right\rbrace $ have to be included in the estimation problem.
	The directed graphs of two possible interpretations are shown in \autoref{fig:GMM} and explained in the following subsections.
	
	\begin{table}[tbph]
		\centering
		\caption{Legend of \autoref{fig:GMM}}
		\label{tab:legend}
		\begin{tabular}{@{}cl@{}}
			\toprule
			\textbf{Symbol}  & \textbf{Name}              \\ \midrule
			$ \info $   & Information Matrix of the GMM Components \\
			$ \mean $   & Mean of the GMM Components        \\
			$ \weight $   & Weight of the GMM Components       \\
			$ \error $   & Measurement Error            \\
			$ \boldsymbol{s} $ & Hidden Variable             \\
			$ \nu $    & Wishart Degree of Freedom        \\
			$ \boldsymbol{V} $ & Wishart Scale Matrix           \\
			$ \beta $   & Information Matrix Scaling        \\
			$N$     & Number of Measurements          \\ \bottomrule
		\end{tabular}
	\end{table}
	
	\subsection{Expectation Maximization}\label{ssec:EM}
	One interpretation, shown in \autoref{fig:GMM_EM}, is to treat $\Params$ as parameters and describe the estimation as maximum-likelihood problem:
	\begin{equation}
	\begin{split}
	&\ParamsOpt = \argmax_{\Params} \prob (\error|\Params) \\
	\text{with } &\prob (\error|\Params) = \int \prob(\error, \boldsymbol{s}|\Params) \mathrm{d}\boldsymbol{s}
	\end{split}
	\label{eqn:em}
	\end{equation}
	Since the integral is intractable, the iterative Expectation-Maximization (EM) algorithm \cite{Dempster1977} splits the problem into two subproblems:
	The E-step estimates the hidden variable $\boldsymbol{s}$ based on an initial guess of $\Params$.
	The M-step estimates the maximum likely set of parameters $\Params$ based on the hidden variables that were estimated before.
	As demonstrated in \cite{Pfeifer2019}, the EM can be applied in an alternating sequence with the least squares optimization of the original state estimation problem \autoref{eqn:max-like}.
	On the downside, the choice of the parameter $K$ is not trivial because it depends on the complexity of the true distribution of $\error$.
	While a too small $K$ results in less robustness, a too large number of components leads to numerical instabilities.
	These instabilities can occur, if only a few samples are responsible for a specific component.
	There exists a variety of EM derivations that aim to overcome this limitation like the greedy EM \cite{Vlassis2002} or the split and merge EM \cite{Ueda2000}, but their parametrization can be even more difficult.
	Even a simple merging criterion is not easy to chose, as the author of \cite{Hennig2010} demonstrates.
	
	\subsection{Bayesian Methods}\label{ssec:bayes}
	A fully Bayesian interpretation of the GMM optimization \autoref{eqn:max-posteriori}, treats the distribution parameters $\Params$ as random variables with corresponding distributions.
	After adding a set of priors $\prob(\Params)$, the estimation problem can be written as:
	\begin{equation}
	\ParamsOpt =\argmax_{\Params} \frac{\prob(\error|\Params) \prob(\Params)}{\int \prob(\error|\Params) \prob(\Params) \mathrm{d}\Params}
	\label{eqn:gmm_bayes}
	\end{equation}
	There exist different suggestions for the choice of this priors.
	In difference to the more common model used in \cite{Blei2006, Rosen2013a, Watson2018a} or \cite[p.475]{Bishop2006}, we follow the suggestion of the authors of \cite{Corduneanu2001} and omit the Dirichlet prior of the mixture's weight $\weight$.
	\autoref{fig:GMM_VBI} shows the corresponding graph with the remaining Wishart prior \autoref{eqn:info_prior} for the information matrix and the normal prior \autoref{eqn:mean_prior} for the mean of the GMM.
	The matrix $\identity$ is the identity matrix here.
	\begin{equation}
	\mean \sim \mathcal{N}(0,\beta_0 \identity)
	\label{eqn:mean_prior}
	\end{equation}
	\begin{equation}
	\info \sim \mathcal{W}(\priorV,\priorNu)
	\label{eqn:info_prior}
	\end{equation}
	To omit the prior of $\weight$ comes with two advantages, one obvious is the reduced number of hyperparameters which makes the algorithm easier to apply.
	The other one is the possibility to reduce weights to zero, which allows us to remove unused components from the model.
	Since the integral in \autoref{eqn:gmm_bayes} can not be calculated directly, approximate solutions like Markov Chain Monte Carlo (MCMC) methods or variational Bayesian inference (VBI) have to be applied.
	MCMC algorithms, such as Gibbs sampling \cite{Geman1984} used in \cite{Rosen2013a} and \cite{Watson2018a}, approximate the integral over the set of parameters $\Params$ by generating many samples of them from their estimated distribution.
	This process is computationally expensive and therefor rather unsuitable for real-time applications, which leads us to variational inference.
	
	The key idea of VBI is to approximate the posterior $\prob(\Params|\error)$ with a distribution $\probVBI(\Params)$.
	\begin{equation}
	\prob(\Params|\error) \approx \probVBI(\Params)
	\label{eqn:vbi}
	\end{equation}
	To find the distributions over all hidden variables $\Params$, it is assumed that they can be partitioned in $j$ independent groups $\left\lbrace \Params_j \right\rbrace$.
	According to these groups, $\probVBI(\Params)$ can be factorized, as shown in \autoref{eqn:vbi_2}, and each factor $\probVBI_{j}(\Params_j)$ can be estimated separately.
	For the model proposed in \cite{Corduneanu2001}, the groups are mean $\mean$, information matrix $\info$ and the correspondence variable $\boldsymbol{s}$.
	\begin{equation}
	\probVBI(\Params) = \prod_{j} \probVBI_{j}(\Params_j) = \probVBI_{\info} (\info) \probVBI_{\mean} (\mean)\probVBI_{\boldsymbol{s}} (\boldsymbol{s})
	\label{eqn:vbi_2}
	\end{equation}
	Since we use exactly the same model that is proposed in \cite{Corduneanu2001}, we refer interested readers to the original work.
	For a more gentle introduction we can also recommend the tutorial paper \cite{Tzikas2008}.
	
	Nevertheless, there are still some drawbacks of Bayesian inference in general and of the concrete VBI solution.
	Bayesian estimators require a set of predefined priors to represent the knowledge that is available in advance.
	If there is little to no prior knowledge about the estimated parameters, it can be hard to specify meaningful priors.
	This work can not provide a general solution for this problem, but we try to find a good trade-off between uninformative priors and prior knowledge that exists about NLOS errors.
	Variational inference require a predefined number of Gaussian components similar to the maximum-likelihood approach.
	However, our approach is robust against an exaggerated number of components, since the weight of unused ones drops to zero without causing singularities.
	
	As a last statement of this section, we want to emphasize that to our knowledge no ``best'' algorithm exists to estimate a Gaussian mixture from empirical data.
	Nevertheless, VBI seems to be well suited for the problem we want to address.
	
	\section{Learning Complexity Online}\label{sec:Complexity}
	
	In this section, we describe the complete algorithm which is composed of robust factor graph optimization and variational Bayesian inference.
	The factor graph applies the GMM to robustly estimate the system's state and the VBI estimates the mixtures based on the empirical distribution of the sensor.
	An overview of the general approach is given by \autoref{fig:Concept} on the first page and an algorithmic description is provided with \autoref{alg:complexity-learning} and \autoref{alg:adpative-mixture}.
	
	As explained in \cite{Pfeifer2019} the estimation problem \autoref{eqn:em-se} can be described as EM algorithm with the set of measurements $\Measurement$ as observed variable and the parameters of the mixture model $\Params$ as hidden ones.
	\begin{equation}
	\begin{split}
	&\StateEst = \argmax_{\State} \prob (\State|\Measurement) \\
	\text{with } &\prob (\State|\Measurement) = \int \prob(\State, \Params|\Measurement) \mathrm{d}\boldsymbol{\Params}
	\end{split}
	\label{eqn:em-se}
	\end{equation} 
	Again, the intractable integral is solved by alternately estimating the expected value of the hidden parameters $\Params$ (E-step) and the state variables (M-step).
	Since the problem is formulated as sliding window estimation, one iteration per time step is enough to achieve good convergence. 
	
	In the E-step, the posterior probability of $\Params$ is calculated with VBI based on a previously determined MAP estimate of the error $\errorEst$:
	\begin{equation}
	\ParamsEst = \argmax_{\Params} \prob (\Params|\errorEstPast)
	\label{eqn:Estep}
	\end{equation}
	The desired expected value $\Expected (\ParamsEst_{t})$ is computed during the variational inference.
	Again, further details are available in \cite{Corduneanu2001}.
	
	The M-step uses the estimated distribution parameter $\Expected (\ParamsEst_{t})$ and applies it to the least squares optimization according to the Sum-Mixture formulation \autoref{eqn:sum_mix}.
	After the optimization, the non-linear error function $\error$ can be used to get a MAP estimate of the error $\error$ as shown in:
	\begin{equation}
	\StateEst_t =\argmax_{\State} \prob(\State|\Measurement_t, \Expected (\ParamsEstNow))
	\label{eqn:Mstep}
	\end{equation}
	\begin{equation}
	\errorEstNow = \error(\StateEst_t,\Measurement_t)
	\label{eqn:error-map}
	\end{equation}
	The complete procedure is described with \autoref{alg:adpative-mixture}.
	
	To adapt the mixture's number of components $K$ to the complexity of the empirical distribution, we add one new component in each time step, starting with $K = 2$ at $t = 0$.
	This leads to a fast rising number of components.
	However, through the variational inference, the weight of unused components drops to zero.
	Therefore, components with a weight below a predefined threshold $\weight_{min}$ can be removed.
	We set this pruning threshold to $\nicefrac{1}{N}$ with $N$ as the number of measurements.
	Each added component is initialized according to its prior distribution with zero mean $\mu_0 = 0$ and the information matrix $\info_0 = \priorNu \priorVinverse$.
	With this simple approach, we are able to choose the right number of components just by ``saturating'' the mixture with components.
	Since the saturation point depends on the parametrization of the Gauss and Wishart prior, a careful parametrization is essential for a good performance. 
	
	To ensure real time capability even for critical applications, we specify an optional upper limit $K_{max}$ for the number of components.
	If this limit is exceeded, the component with the lowest weight is removed before a new one is added.
	The online complexity learning is described by \autoref{alg:complexity-learning}.
	
	\begin{algorithm}[tbph]
		\SetAlgoLined
		\KwData{$\errorEstNow$, $\ParamsEstPast$}
		\KwResult{$\ParamsEstNow$}
		
		\Begin(Initialization of $\ParamsEstNow$) 
		{
			$\ParamsEstNow$ = $\ParamsEstPast$\;
			
			\If{$K \geq K_{max}$}
			{
				Remove the smallest component from $\ParamsEstNow$\;
			}
			
			Add new component to $\ParamsEstNow$\;
		}
		
		\Begin(Estimation of $\ParamsEstNow$) 
		{
			\Repeat{$i > i_{max}$ or $\Delta L < \Delta L_{min}$}
			{
				Perform one iteration of VBI \cite{Corduneanu2001}\;
				
				\If{any $\weight_k < \weight_{min}$}
				{
					Remove component k from $\ParamsEstNow$\;
				}
				
				Increment iteration count $i$\;
				
				Compute relative likelihood $\Delta L$\;
			}
		}
		\caption{Complexity Learning (CL)}
		\label{alg:complexity-learning}
	\end{algorithm}
	
	\begin{algorithm}[tbph]
		\SetAlgoLined
		\KwData{$\Measurement$}
		\KwResult{$\StateEst$}
		
		Initially estimate $\StateEst_{t=0}$ with \autoref{eqn:least_squares}\;
		
		Perform \autoref{alg:complexity-learning} twice to get $\ParamsEst_{t=0}$\;
		
		\ForEach{time step $t$}
		{
			Add $\measurement_t$, $\state_t$ to $\prob(\Measurement|\State)$\;
			
			Remove $\measurement$, $\state$ older than \SI{60}{\second} from $\prob(\Measurement|\State)$\;
			
			Calculate $\errorEstNow$\;
			
			Perform \autoref{alg:complexity-learning} to get $\ParamsEstNow$\;
			
			Update $\prob(\Measurement|\State)$ with $\ParamsEstNow$\;
			
			Estimate $\StateEst_t$ by optimizing $\prob(\Measurement|\State)$\;
			
		}
		
		\caption{Incremental Variational Mixture (IVM)}
		\label{alg:adpative-mixture}
	\end{algorithm}
	
	\section{Evaluation}\label{sec:Eval}
	\begin{table*}[tbph]
		\centering
		\caption{Accuracy and Runtime of the Final Evaluation.}
		\label{tab:result}
		\setlength\tabcolsep{6pt}
		\begin{tabular}{@{}lllllllllll@{}}
			\toprule
			& \multicolumn{2}{c}{Chemnitz} & \multicolumn{2}{c}{Berlin PP} & \multicolumn{2}{c}{Berlin GM} & \multicolumn{2}{c}{Frankfurt MT} & \multicolumn{2}{c}{Frankfurt WT} \\
			\multirow{-2}{*}{Algorithm} & ATE [m] & Time [s] & ATE [m] & Time [s] & ATE [m] & Time [s] & ATE [m] & Time [s] & ATE [m] & Time [s] \\ \midrule
			Gaussian & 30.0 & 58.6 & 29.2 & 9.5 & 13.38 & 49.7 & 30.97 & 58.1 & 23.54 & 31.9 \\
			DCS \cite{Agarwal2013}& 4.403 & 54.9 & 25.04 & 14.7 & 19.11 & 64.5 & 13.25 & 69.5 & 11.39 & 35.2 \\
			cDCE \cite{Pfeifer2017} & 4.326 & 54.5 & 17.91 & 14.3 & 14.59 & 65.3 & 14.93 & 73.6 & 11.12 & 34.9 \\
			SM+EM \cite{Pfeifer2019} & \bestResult{2.378} & 102.0 & 12.45 & 39.8 & 13.88 & 163.0 & 10.72 & 136.0 & 6.42 & 76.5 \\
			SM+VBI & 3.106 & 119.0 & 12.4 & 33.3 & 13.55 & 186.0 & 11.23 & 131.0 & \bestResult{4.117} & 84.4 \\
			SM+EM+CL & 6.723 & 172.0 & 12.23 & 45.0 & 13.37 & 312.0 & 10.41 & 214.0 & 9.706 & 152.0 \\
			IVM (SM+VBI+CL) & 2.48 & 122.0 & \bestResult{11.56} & 59.4 & \bestResult{10.99} & 352.0 & \bestResult{8.586} & 248.0 & 4.626 & 146.0 \\
			\bottomrule
		\end{tabular}
	\end{table*}
	\begin{figure*}[tbph]
		\centering
		\includegraphics[width=\linewidth]{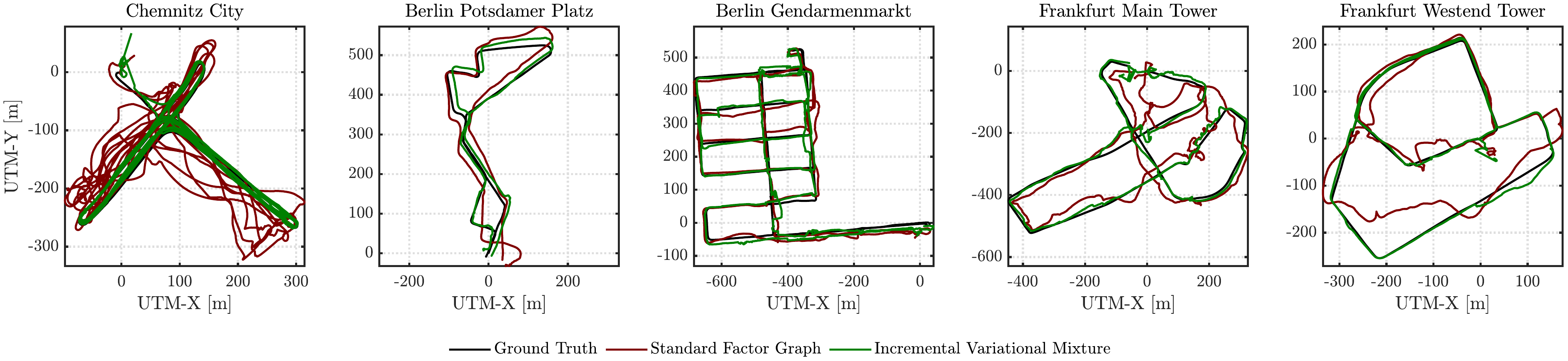}
		\caption{Resulting UTM trajectories of the non-robust factor graph (red) and the proposed Incremental Variational Mixture algorithm (green). The distortions caused by NLOS are significantly reduced.}
		\label{fig:trajectoryiv}
	\end{figure*}
	
	In this section, we demonstrate the performance of the proposed Incremental Variational Mixture (IVM) algorithm on several real world GNSS datasets.
	The resulting trajectories are shown in \autoref{fig:trajectoryiv}.
	We do not only compare against the robust state-of-the-art approaches DCS \cite{Agarwal2013} and cDEC \cite{Pfeifer2017} but also show the impact of VBI and the proposed Complexity Learning (CL) mechanism.
	Beside a comparison based on position error metrics, we want to demonstrate the sensitivity of our approach to its parametrization, which is important for practical applications.
	To proof the performance under real time conditions, we perform the estimation under online conditions.
	This means, results are calculated and stored without information from future measurements.
	
	\subsection{The Datasets}
	We use five different datasets, that consist of raw pseudorange measurements from a mass market receiver and wheel odometry from measurement vehicle.
	These datasets are collected in different urban areas in Germany and have an accumulated length of \SI{18.6}{\kilo\meter} respectively \SI{92}{\minute}.
	A precise ground truth is provided by a combination of differential GNSS and a tactical grade inertial measurement unit.
	Technical details are published in \cite{Reisdorf2016} and the datasets themselves are available online \footnote{\url{http://mytuc.org/GNSS}}.
	
	\subsection{Parametrization}
	Since the choice of parameters has a significant impact on the performance of the proposed algorithms, we want to explain them clearly.
	The noise parameters of the factor graph are identical to our previous work, therefore we want to refer to \cite{Pfeifer2019} for detailed information.
	\begin{table}[htbp]
		\centering
		\caption{Parameters of the Incremental Variational Mixture Algorithm}
		\label{tab:params_vbi}
		\begin{tabular}{@{}llc@{}l}
			\toprule
			\textbf{Parameter Name}         & \textbf{Symbol}  &                 \textbf{Value}                  &  \\ \midrule
			Wishart Prior Degree of Freedom & $\priorNu$       &                        2                        &  \\ \midrule
			Wishart Prior Scale Matrix      & $\priorV$        & $\nicefrac{\var\left( \errorEst \right) }{\nu}$ &  \\ \midrule
			Normal Prior Information Matrix & $\beta_0$        &         $\SI{e-6}{\per\meter\squared}$          &  \\ \midrule
			Pruning Threshold               & $\weight_{min}$  &                $\nicefrac{1}{N}$                &  \\ \midrule
			Max. Iterations                 & $i_{max}$        &                     $1000$                      &  \\ \midrule
			Min. $\Delta$Likelihood         & $\Delta L_{min}$ &                    $10^{-6}$                    &  \\ \midrule
			Max. Number of Components       & $K_{max}$        &                       $8$                       &  \\ \bottomrule
		\end{tabular}
	\end{table}
	
	To apply the variational inference, several parameters have to be specified that are summarized in \autoref{tab:params_vbi}.
	Not all of them are equally important, especially the convergence criteria and the pruning threshold $\weight_{min}$ have just a small impact on the overall performance.
	The optional maximum of Gaussian components $K_{max}$ is also easy to parametrize, it has to be high enough to reflect the worst case complexity of the empirical distribution.
	A value of $K_{max} = 8 \dots 12$ should be enough for all practical applications and if the algorithm is applied to a fully unknown problem, the limit can be omitted.
	The information matrix of the normal prior $\beta_0$ defines the initial uncertainty about each component's mean.
	Since NLOS errors over \SI{1}{\kilo\meter} are very unlikely, a value of $\beta_0 = \nicefrac{1}{(\SI{1}{\kilo\meter})^2}$ seems to be sufficient.
	
\begin{figure}[tbp]
		\centering
		\includegraphics[width=\linewidth]{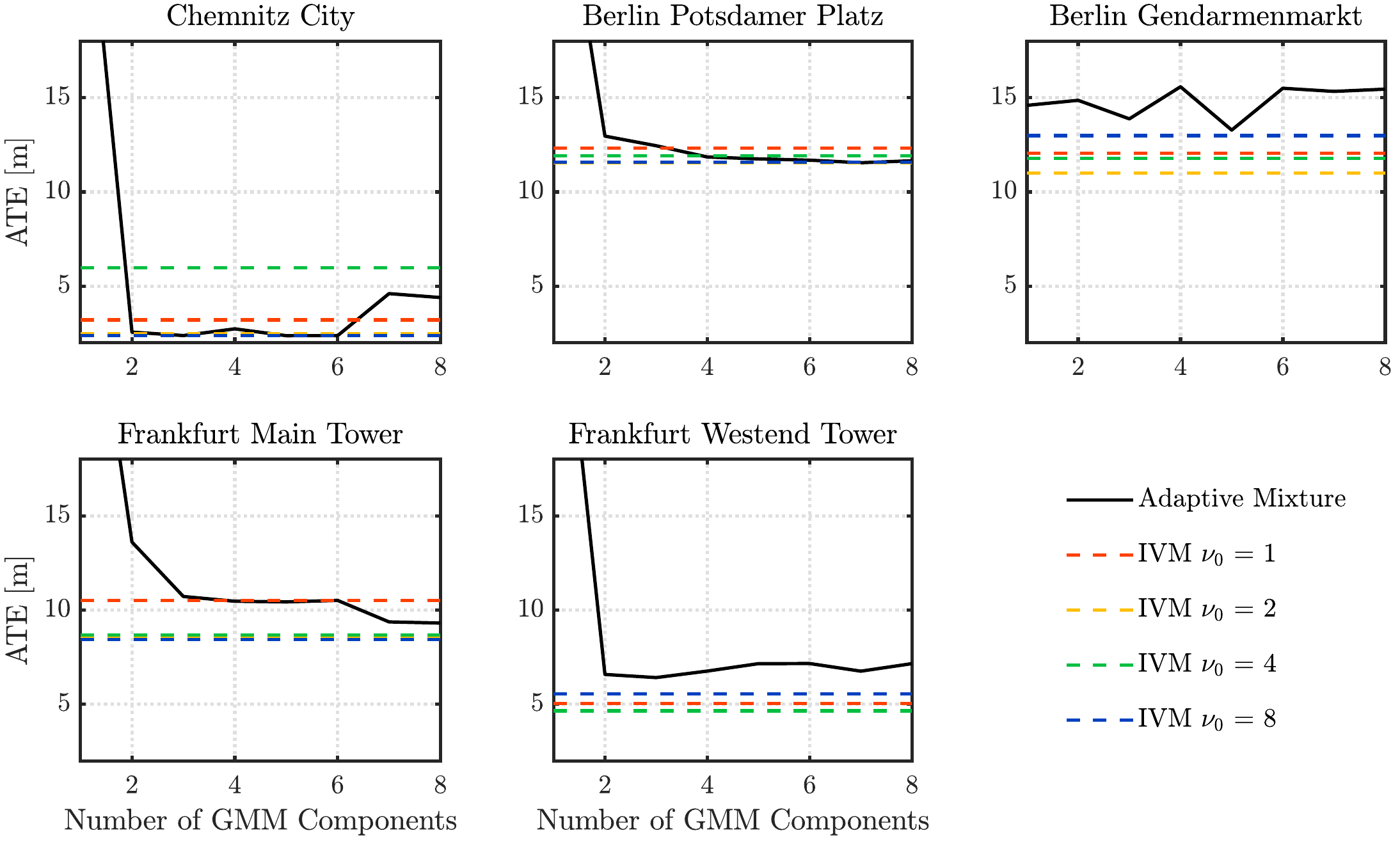}
		\caption{
			Comparison between the Adaptive Mixture approach \cite{Pfeifer2019} and the proposed Incremental Variational Mixture (IVM) algorithm on all datasets.
			The solid black line shows the mean ATE for different choices of the number of mixture components.
			Each dotted line corresponds to a run of the IVM with a specific $\priorNu$.
			Since the number of components is variable here, we draw it as horizontal line.
			The comparison shows that IVM outperforms the other algorithm for a $\priorNu$ of 2.
		}
		\label{fig:ParamEval}
	\end{figure}

	More difficult is the parametrization of the Wishart prior $\mathcal{W}(\priorV,\priorNu)$.
	The scale matrix $\priorV$ in combination with the degree of freedom $\priorNu$ determines the expected information matrix of the estimated mixture components:
	\begin{equation}
	\Expected(\info) = \priorNu \cdot \priorVinverse
	\label{eqn:wishart_v}
	\end{equation} 
	A coarse guess of the information matrix can be estimated directly from the error $\errorEst$ by calculating its variance.
	So, a meaningful prior can be defined with:
	\begin{equation}
	\priorV = \frac{\var(\errorEst)}{\priorNu}
	\label{eqn:wishart_v2}
	\end{equation}
	For the positive integer $\priorNu$, a straightforward approximation is not possible.
	The authors of the variational model \cite{Corduneanu2001} recommend an informative Wishart prior, which means value of $\priorNu = 1$ or at least close to 1.
	We evaluated the proposed algorithm with different values and visualized the localization error in \autoref{fig:ParamEval}.
	Therefore, we chose a prior with $\priorNu = 2$ to achieve the best performance over all datasets.
	The reason for the strong impact of $\priorNu$ to the overall result is the convergence of the VBI algorithm.
	An increased $\priorNu$ improves the convergence speed and increases the number of removed mixture components.
	With $\priorNu = 1$, the number of components over-fits the empirical distribution which leads to more local minima in the state estimation problem and therefor reduced accuracy.
	Although, the performance of different values of $\priorNu$ is not fully consistent over all datasets, the choice of $\priorNu = 2$ generalizes at least for the tested ones.
	The comparison to the EM based algorithm in \autoref{fig:ParamEval} shows that its fixed number of components $K$ also have to be chosen carefully.
	We set $K = 3$ for a fair comparison since it is the best choice over all datasets.
	
	\subsection{Implementation Details}
	We implemented the algorithm as part of our robust sensor fusion library libRSF\footnote{\url{http://mytuc.org/libRSF}}.
	The VBI algorithm will be published as open source in a future release.
	The non-linear least squares optimization is based on the Ceres solver \cite{Agarwal} and the GMM estimation is implemented in C++ using Eigen \cite{Guennebaud2010} without multi-threading.
	All tests were performed on a Intel i7-7700 system.
	
	\subsection{Results}
	Metric of our comparison is the absolute trajectory error (ATE).
	We define the ATE as euclidean distance between estimated trajectory and ground truth in the local XY-plane.
	\autoref{tab:result} summarizes the mean ATE as well as the runtime of the evaluated algorithms.
	Along with the proposed IVM, we compare different combinations of self-tuning algorithms.
	Common is the Sum-Mixture (SM) method to represent a GMM, but we differentiate between EM and VBI estimation and whether the proposed Complexity Learning (CL) is applied or not.
	Please notice that SM+EM equals the Adaptive Mixture approach from \cite{Pfeifer2019} and SM+VBI+CL is the proposed IVM algorithm.
	
	Both VBI and CL seem to have a positive influence regarding the localization accuracy.
	Although, only in combination, they are able to consistently reduce the ATE over all datasets.
	Therefore, we would prefer the proposed IVM for applications where a maximum precision is required.
	Drawback of the CL is increased runtime of the algorithm, nevertheless it is 5 times faster than the recording time of the datasets.
	
	If runtime is more important, we would tend to prefer the Adaptive Mixture (SM+EM) approach, since it offers a solid performance without the trouble of parameterizing the variational priors.
	
	The results of the M-estimators DCS and cDCE show improvements compared to the non-robust factor graph, but both fall behind the GMM based methods.
	
	\section{Conclusion}\label{sec:Conclusion}
	The proposed algorithmic approach, allows to apply least squares optimization to sensor fusion problems with non-Gaussian and time dependent error distributions like GNSS localization in urban environments.
	We not only adapt the parameters of the Gaussian mixture representation, we also learn the right number of components to represent its complexity.
	The applied variational Bayesian inference allows a numerically stable and real time capable solution.
	
	The comparison against several state-of-the-art algorithms were performed on a set of open access GNSS datasets that reflect the localization of an autonomous vehicle in an urban environment.
	Our approach demonstrated a superior estimation quality over the majority of datasets in combination with an increased but acceptable computation time.
	
	The described connection between state estimation and adaptive error models opens a wide field of possible future improvements.
	Since the parametrization of the variational priors is the most critical drawback currently, we will address it in future work.
	Variational Bayesian inference also offers further methods to adaptively split and merge Gaussian components which could be explored in this context.
	Finally, a broader evaluation on different domains like SLAM would be useful to validate the approach and the choice of parameters.
	
	\footnotesize
	\bibliographystyle{IEEEtranN}
	\bibliography{IEEEabrv,CleanBib}
\end{document}

%% file: math.tex
\DeclareMathOperator*{\argmin}{\arg\!\min}
\DeclareMathOperator*{\argmax}{\arg\!\max}

\DeclareMathOperator{\var}{\operatorname{var}}

\newcommand{\vect}[1]{\mathbf{#1}}
\newcommand{\est}[1]{\hat{#1}}
\newcommand{\opt}[1]{#1^\ast}

\newcommand{\State}{\vect{X}}
\newcommand{\StateOpt}{\opt{\vect{X}}}
\newcommand{\StateEst}{\vect{\est{X}}}

\DeclareMathOperator{\identity}{\boldsymbol{I}}

\newcommand{\state}{\vect{x}}
\newcommand{\stateOf}[2]{\state^{#1}_{#2}}

\newcommand{\statePose}{\stateOf{x,y,z,\phi}{t}}
\newcommand{\statePoseFuture}{\stateOf{x,y,z,\phi}{t+1}}
\newcommand{\statePseudorange}{\stateOf{x,y,z,\delta}{t}}

\newcommand{\stateCCED}{\stateOf{\delta,\dot{\delta}}{t}}
\newcommand{\stateCCEDFuture}{\stateOf{\delta,\dot{\delta}}{t+1}}

\newcommand{\Measurement}{\vect{Z}}

\newcommand{\measurement}{\vect{z}}
\newcommand{\measurementOf}[2]{\measurement^{#1}_{#2}}

\newcommand{\measurementPseudorange}{\measurementOf{pr}{t,i}}

\newcommand{\measurementOdom}{\measurementOf{odo}{t}}

\newcommand{\error}{\vect{e}}
\newcommand{\errorOf}[2]{\error^{#1}_{#2}}

\newcommand{\errorEst}{\vect{\est{e}}}
\newcommand{\errorEstNow}{\vect{\est{e}}_t}
\newcommand{\errorEstPast}{\vect{\est{e}}_{t-1}}

\newcommand{\errorPseudorange}{\errorOf{pr}{t,i}}

\newcommand{\prob}{\vect{P}}

\DeclareMathOperator{\Expected}{\mathbb{E}}

\newcommand{\mean}{{\boldsymbol{\mu}}}
\newcommand{\weight}{w}
\newcommand{\info}{\boldsymbol{\mathcal{I}}}

\newcommand{\sqrtinfoOf}[1]{\info_{#1}^{\frac{1}{2}}}

\newcommand{\sqrtinfok}{\sqrtinfoOf{k}}
\newcommand{\sqrtinfon}{\sqrtinfoOf{n}}

\newcommand{\Params}{\boldsymbol{\theta}}
\newcommand{\ParamsOpt}{\opt{\Params}}
\newcommand{\ParamsEst}{\boldsymbol{\est{\theta}}}
\newcommand{\ParamsEstNow}{\ParamsEst_t}
\newcommand{\ParamsEstPast}{\ParamsEst_{t-1}}

\DeclareMathOperator{\probVBI}{\mathbf{Q}}

\DeclareMathOperator{\priorNu}{\nu_0}
\DeclareMathOperator{\priorV}{\boldsymbol{V}_0}
\DeclareMathOperator{\priorVinverse}{\boldsymbol{V}_{0}^{-1}}